\documentclass[letterpaper, 10 pt, conference]{ieeeconf}  

\IEEEoverridecommandlockouts                              

\usepackage{amsmath,amsfonts}
\usepackage{amssymb}
\usepackage{algorithmic}
\usepackage{array}
\usepackage{caption}
\usepackage[caption=false,font=normalsize,labelfont=sf,textfont=sf]{subfig}
\usepackage{textcomp}
\usepackage{stfloats}
\usepackage{url}
\usepackage{hyperref}
\usepackage{verbatim}
\usepackage{graphicx}
\usepackage{xcolor,colortbl}
\usepackage{tabularx,booktabs}
\usepackage{array,multirow,graphicx}
\usepackage{makecell}
\usepackage[export]{adjustbox}
\usepackage{pifont} 
\usepackage{etoolbox}
\usepackage[normalem]{ulem}

\let\labelindent\relax
\usepackage{enumitem} 
\usepackage[dvipsnames]{xcolor}

\colorlet{colorFst}{Green!25}       %
\colorlet{colorSnd}{SpringGreen!45} %
\colorlet{colorTrd}{Yellow!30}      %
\newcommand{\fs}{\cellcolor{colorFst}\bf}   
\newcommand{\nd}{\cellcolor{colorSnd}}      
\newcommand{\rd}{\cellcolor{colorTrd}}      

\definecolor{pink}{RGB}{255,105,180}
\hypersetup{
    urlcolor=pink
}

\title{\LARGE \bf
MCGS-SLAM: A Multi-Camera SLAM Framework Using Gaussian Splatting for High-Fidelity Mapping
}

\author{Zhihao Cao$^{1}$, Hanyu Wu$^{2}$, Li Wa Tang$^{2}$, Zizhou Luo$^{3}$, \\Wei Zhang$^{4}$, Marc Pollefeys$^{5, 6}$, Zihan Zhu$^{5, *}$, and Martin R. Oswald$^{7}$
\thanks{$^{*}$Zihan Zhu is the Project Lead of this work.}
\thanks{$^{1}$Zhihao Cao is with the Department of Mathematics, 
        ETH Zurich, Switzerland. (e-mail: zhicao@student.ethz.ch)}%
\thanks{$^{2}$Hanyu Wu and Li Wa Tang are with the Department of Mechanical and Process Engineering, 
        ETH Zurich, Switzerland. (e-mail: hanywu@student.ethz.ch; litang1@student.ethz.ch)}%
\thanks{$^{3}$Zizhou Luo is with the Department of Informatics, 
        University of Zurich, Switzerland. (e-mail: zizhou.luo@uzh.ch)}%
\thanks{$^{4}$Wei Zhang is with the Institute for Photogrammetry, University of Stuttgart, Germany (e-mail: wei.zhang@ifp.uni-stuttgart.de)}%
\thanks{$^{5}$Marc Pollefeys and Zihan Zhu are with Computer
Vision and Geometry Group, ETH Zurich, 8092 Zurich, Switzerland. (e-mail: zihan.zhu@inf.ethz.ch; marc.pollefeys@inf.ethz.ch)}%
\thanks{$^{6}$Marc Pollefeys is also with Microsoft Spatial AI Lab, 8038
Zurich, Switzerland (e-mail: mapoll@microsoft.com)}%
\thanks{$^{7}$Martin R. Oswald is with Computer Vision Research Group, University of Amsterdam, Netherlands (e-mail: m.r.oswald@uva.nl)}%
}

\begin{document}

\newcommand{\acceptednote}{
\begin{center}
\vspace{-1.0cm}
\small
{\color{gray}This paper has been accepted for publication at the IEEE International Conference on Robotics and Automation (ICRA), 2026~\copyright~IEEE}
\end{center}
\vspace{-0.4cm}
}

\newcommand{\insertfig}{
   \vspace{0.4cm}
   \captionsetup{type=figure}
   \includegraphics[width=\textwidth]{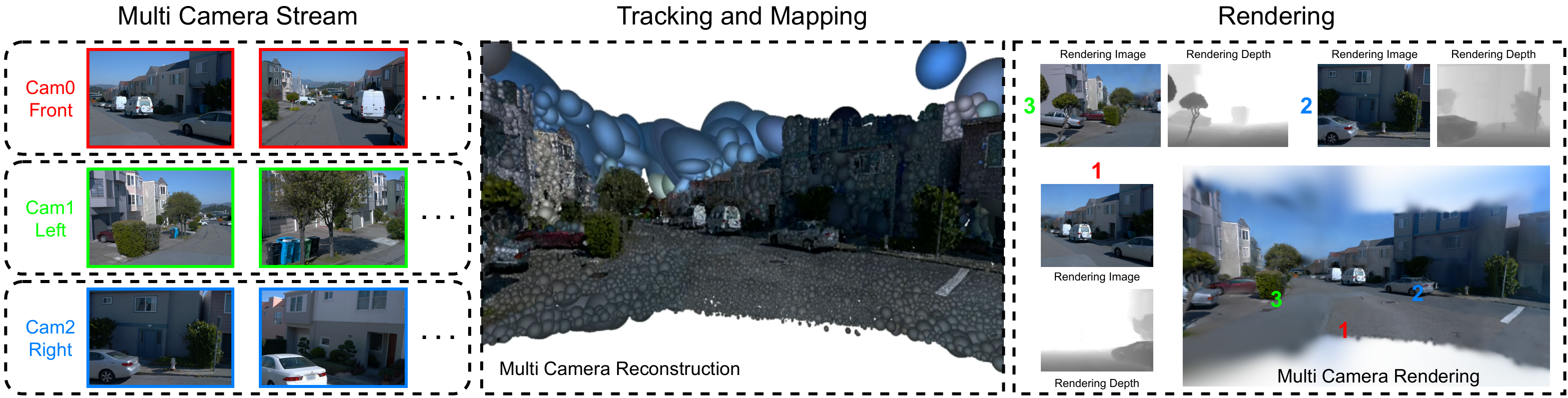}  
   \vspace{-0.3cm} 
   \setcounter{figure}{0}
   \captionof{figure}{MCGS-SLAM synchronizes RGB inputs from the front, left, and right cameras of the multi-camera rig in the Waymo dataset and fuses them into a unified 3D Gaussian Splatting map. The system performs real-time tracking and mapping, enabling high-fidelity reconstruction of both color and depth views from each individual camera. Through joint multi-camera optimization, MCGS-SLAM ensures accurate pose and geometry alignment, while supporting comprehensive multi-view rendering for photorealistic visualization. \url{https://zhihao-ethz.github.io/mcgs-slam/}}
  \label{fig:teaser}
  \vspace{-0.15cm} 
}

\makeatletter
\pretocmd{\@maketitle}{\acceptednote}{}{}
\apptocmd{\@maketitle}{\insertfig}{}{}



\maketitle
\pagestyle{empty}

\begin{abstract}

Recent progress in dense SLAM has primarily targeted monocular setups, often at the expense of robustness and geometric coverage. We present MCGS-SLAM, the first purely RGB-based multi-camera SLAM system built on 3D Gaussian Splatting (3DGS). Unlike prior methods relying on sparse maps or inertial data, MCGS-SLAM fuses dense RGB inputs from multiple viewpoints into a unified, continuously optimized Gaussian map. A multi-camera bundle adjustment (MCBA) jointly refines poses and depths via dense photometric and geometric residuals, while a scale consistency module enforces metric alignment across views using low-rank priors. The system supports RGB input and maintains real-time performance at large scale. Experiments on synthetic and real-world datasets show that MCGS-SLAM consistently yields accurate trajectories and photorealistic reconstructions, usually outperforming monocular baselines. Notably, the wide field of view from multi-camera input enables reconstruction of side-view regions that monocular setups miss, critical for safe autonomous operation. These results highlight the promise of multi-camera Gaussian Splatting SLAM for high-fidelity mapping in robotics and autonomous driving.

\end{abstract}


\section{Introduction}
\label{sec:intro}

Simultaneous Localization and Mapping (SLAM) remains a foundational component in robotic navigation and 3D scene reconstruction. Early monocular SLAM systems, such as ORB-SLAM~\cite{mur2015orb, campos2021orb}, LSD-SLAM~\cite{engel2014lsd}, and DSO~\cite{wang2017stereo}, achieve real-time camera tracking by minimizing sparse geometric or photometric residuals. However, their reliance on a single narrow field-of-view (FoV) camera renders them susceptible to scale drift, motion blur, and occlusions. Learning-augmented approaches, including DROID-SLAM~\cite{teed2021droid} and MAC-VO~\cite{qiu2024mac}, alleviate some of these issues, yet the core limitation of monocular viewpoint remains a fundamental bottleneck for scene completeness and depth accuracy. A natural solution is to employ overlapping multi-camera systems. Early visual-inertial odometry pipelines improved robustness through fisheye clusters~\cite{urban2016multicol}. More recently, Kuo et al.~\cite{kuo2020redesigning} proposed a generalization of visual-inertial bundle adjustment (BA) to wide-baseline multi-camera systems through adaptive initialization and keyframe selection. BAMF-SLAM~\cite{zhang2023bamf} introduced a scalable BA formulation for general camera networks, achieving state-of-the-art odometry accuracy. Nevertheless, these systems typically yield only sparse landmarks, and some systems heavily rely on inertial sensors, relegating high-fidelity geometry and photorealistic rendering to costly offline post-processing.

In parallel, dense scene representations have made remarkable strides, though predominantly in monocular settings, thus underutilizing the potential of multi-camera platforms. Traditional map structures such as surfels and TSDF volumes~\cite{newcombe2011kinectfusion, kerl2013dense} have evolved towards neural implicit fields. NeRF-based methods~\cite{mildenhall2021nerf, muller2022instant} enable impressive photorealism, while SLAM variants like NICER-SLAM~\cite{zhu2024nicer} and GLORIE-SLAM~\cite{zhang2024glorie} integrate neural fields into SLAM pipelines for high-quality novel view synthesis. However, these methods remain computationally expensive and lack explicit geometric control. In contrast, 3D Gaussian Splatting (3DGS) \cite{kerbl20233d} offers an efficient alternative that combines explicit geometry, differentiable rasterization, and fast optimization. Recent extensions, including MonoGS~\cite{matsuki2024gaussian} for dense tracking, Loop-Splat~\cite{zhu2024loopsplat} for loop closure, Splat-SLAM~\cite{sandstrom2025splat} for global joint optimization, and HI-SLAM2~\cite{zhang2024hi} for monocular refinement, demonstrate strong results. Still, they inherit the limitations of monocular input: limited FoV, scale ambiguity, and degraded performance in low-texture or occluded regions. These drawbacks highlight the unmet potential of fusing multi-view observations with the efficiency of Gaussian splatting. While multi-agent extensions~\cite{yugay2025magic} also support multiple cameras, they cannot benefit from calibrated rigs.

Leveraging a calibrated multi-camera rig with $k$ spatially overlapping views offers rich observational redundancy but presents challenges in fusing dense RGB streams into a unified Gaussian representation, specifically, maintaining inter-camera scale consistency, achieving drift-free tracking, and enabling efficient online mapping with large numbers of Gaussians. We propose MCGS-SLAM, to the best of our knowledge, the first fully vision-based multi-camera SLAM system built upon 3D Gaussian Splatting with purely RGB input. MCGS-SLAM jointly estimates accurate camera trajectories and high-fidelity 3D reconstructions by fusing raw RGB inputs into a globally consistent Gaussian map. Our framework also supports RGB-D inputs, but this paper focuses on the RGB-only setting. Central to our framework is a Multi-Camera Bundle Adjustment (MCBA) module that jointly optimizes pose and dense depth across views via photometric and geometric consistency. To ensure metric-scale alignment, we introduce a complementary module that leverages low-rank geometric priors from a learned network. These components enable scalable Gaussian optimization and pruning across large anisotropic fields, yielding reconstructions with sharp geometry and photorealistic textures under wide baselines. Our contributions are as follows.
\begin{itemize}[itemsep=0pt,topsep=2pt,leftmargin=10pt]
\item An efficient multi-camera Gaussian SLAM system supporting RGB inputs, with joint optimization over camera poses and 3DGS maps.
\item A unified multi-camera framework that combines Multi-Camera Bundle Adjustment (MCBA) and Joint Depth–Scale Alignment (JDSA), jointly optimizing photometric consistency, geometric priors, and global scale alignment across views.
\item A practical and scalable implementation that generalizes across real-world and synthetic benchmarks, demonstrating strong performance in both geometry and appearance.
\end{itemize}

Through these innovations, MCGS-SLAM bridges the gap between wide-baseline multi-camera tracking and dense 3D Gaussian mapping, laying the groundwork for next-generation robotic perception, digital twin construction, and autonomous systems at scale.
\begin{figure}[tb]
  \centering
  \includegraphics[scale=0.22]{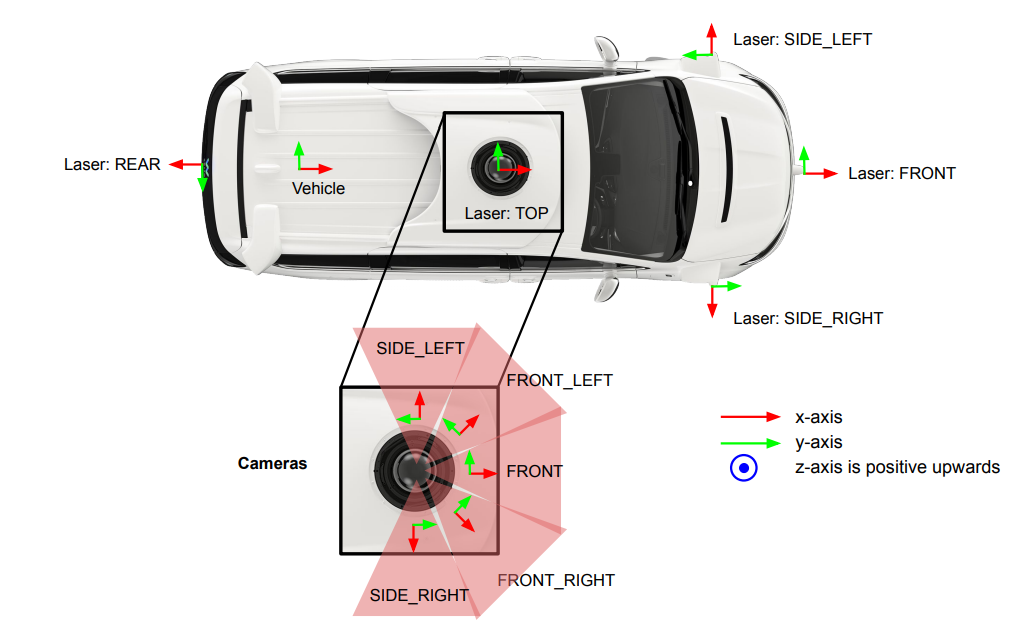}
  \vspace{-0.3cm}
  \caption{The sensor suite integrates multiple wide-angle RGB cameras centrally mounted on the vehicle’s roof in Waymo Open Dataset \cite{sun2020scalability}, whose fan-shaped fields of view collectively provide full $240^\circ$ coverage. This configuration enables high-density observations for multi-camera SLAM and autonomous driving algorithms.}
  \label{fig:waymo_dataset}
  \vspace{-0.6cm}
\end{figure}

\section{Preliminaries}
\label{sec:prelim}

This section introduces the core concepts underpinning our multi-camera Gaussian Splatting SLAM framework. We first review the dense SLAM formulation and the multi-camera setting, followed by an overview of Recurrent Field Transforms and learning-based SLAM front-ends such as DROID-SLAM and BAMF-SLAM. Finally, we present the 3D Gaussian Splatting representation, which serves as the foundational structure of our mapping system.

\subsection{Problem Setting: Dense Multi-Camera SLAM}

\subsubsection{Dense SLAM}
Given a temporally ordered stream of color (or color–depth) images ${I_t}$ captured at time $t$ by a calibrated rig with $k$ camera views, dense SLAM jointly estimates the metric camera trajectory $\mathbf{T} = \{\mathbf{T}_t\}_{t=0}^{L}$ with $\mathbf{T}_t \in \mathrm{SE(3)}$ and a continuous scene map $\boldsymbol{\mathcal{M}}$ by minimizing photometric and geometric residuals across all pixels. To ensure both temporal and spatial consistency, we define a set of frame pairs $(t,t') \in \mathcal{E}$, where $t'$ denotes either a temporally adjacent frame or one selected via keyframe heuristics. The overall objective is formulated as
\begin{equation}
  \mathop{\arg\min}_{\mathbf{T},\boldsymbol{\mathcal{M}}}\;
  \sum_{(t,t')\in\mathcal{E}}\,
  \Bigl\|
     I_{t}-I_{t'}\circ
     \Pi\!\bigl(\mathbf{T}_{tt'}\,\Pi^{-1}( \mathbf{p}_{t},d_{t})\bigr)
  \Bigr\|_{\rho},
  \label{eq:dense_ba}
\end{equation}
where $\Pi$ and $\Pi^{-1}$ denote the projection and back-projection functions, $d_t$ is the depth at pixel $\mathbf{p}_t$, and $\rho(\cdot)$ is the robust $\ell_2$ penalty function. The transformation $\mathbf{T}_{tt'}$ denotes the relative camera pose from frame $t$ to $t'$. Unlike pipelines based on sparse features, optimization in \eqref{eq:dense_ba} is performed at full image resolution, enabling recovery of dense scene geometry.

\subsubsection{Multi-Camera Setting}
The calibrated multi-camera system is defined by fixed extrinsic transformations $\mathbf{T}_{C}^{B} \in \mathrm{SE(3)}$, which map points from each individual camera frame $C$ to a shared body frame $B$. As illustrated in Fig.~\ref{fig:waymo_dataset}, modern automotive datasets such as the Waymo Open Dataset provide time-synchronized, wide-baseline camera clusters composed of multiple global-shutter RGB sensors with accurate intrinsic and extrinsic calibrations. These offer a compelling testbed for SLAM systems, as they introduce strong parallax, large fields of view, and a complex environment.

\begin{figure*}[htpb]
  \centering
  \includegraphics[scale=0.305]{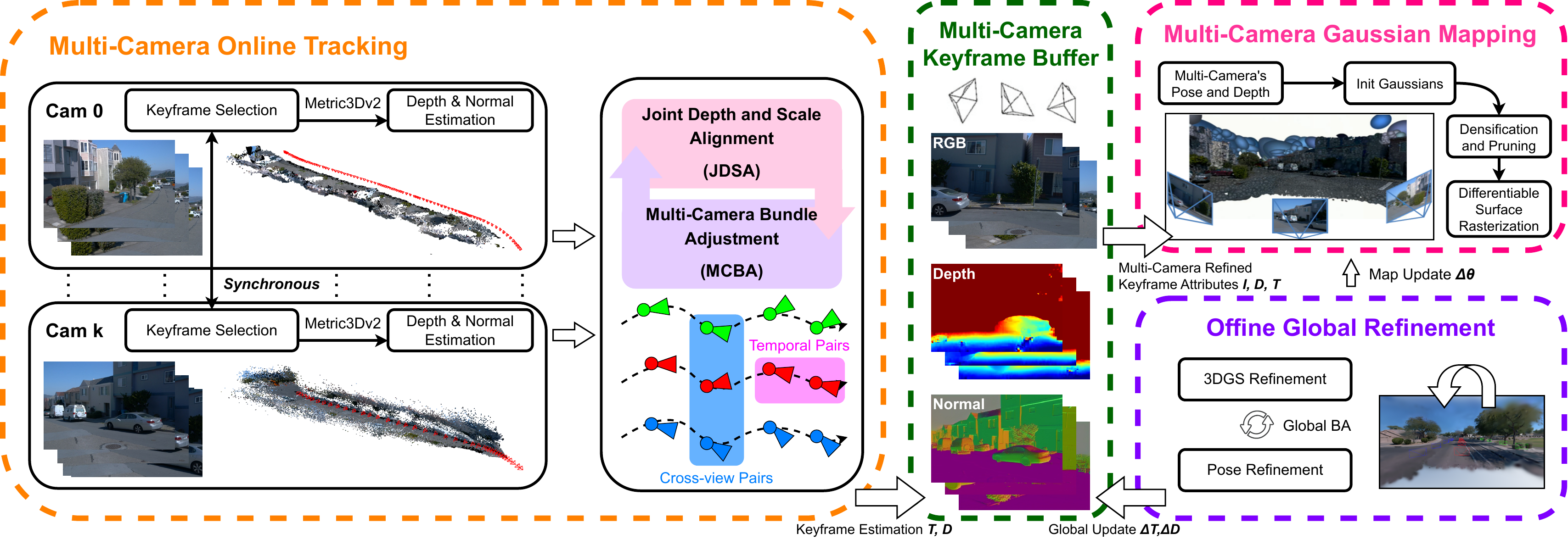}
  \vspace{-0.1cm} 
  \caption{Our method performs real-time SLAM by fusing synchronized inputs from a multi-camera rig into a unified 3D Gaussian map. It first selects keyframes and estimates depth and normal maps for each camera, then jointly optimizes poses and depths via multi-camera bundle adjustment and scale-consistent depth alignment. Refined keyframes are fused into a dense Gaussian map using differentiable rasterization, interleaved with densification and pruning. An optional offline stage further refines camera trajectories and map quality. The system supports RGB inputs, enabling accurate tracking and photorealistic reconstruction.}
  \label{fig:method_overview}
  \vspace{-0.5cm}  
\end{figure*}

\subsection{Recurrent Field Transforms and Learning-based SLAM}

Recurrent Field Transforms (RFT) extend the RAFT family of recurrent optical flow networks to iteratively refine dense correspondences between two views. Given a current reprojection $\hat{\mathbf{p}}_{ij}$, RFT predicts a flow increment $\boldsymbol{\delta}_{ij}$ and an associated per-pixel confidence weight $w_{ij}$. The refined target location is defined as $\tilde{\mathbf{p}}_{ij} = \hat{\mathbf{p}}_{ij} + \boldsymbol{\delta}_{ij}$ and is used to minimize the reprojection error. During optimization, the resulting weighted residual is inserted into the normal equations of bundle adjustment (BA) \cite{teed2021droid} as
\begin{equation}
  r_{ij}=
  \bigl\|\tilde{\mathbf{p}}_{ij} -
  \Pi\!\bigl(\mathbf{T}_{ij}\Pi^{-1}(\mathbf{p}_{i},d_{i})\bigr)\bigr\|_{w_{ij}}^2
  \label{eq:rft_residual}
\end{equation}
where $\Pi$ and $\Pi^{-1}$ denote projection and back-projection, respectively. Equation~\eqref{eq:rft_residual} forms the foundation of the dense, differentiable front-end in DROID-SLAM. To tightly couple correspondence estimation and geometric optimization, DROID-SLAM augments classical photometric BA with RFT, treating optical flow as a latent variable updated via a gated recurrent unit (GRU). This formulation enables joint, real-time optimization of camera poses, per-frame depths, and inter-frame flow, achieving state-of-the-art accuracy in monocular visual odometry. BAMF-SLAM builds upon DROID-SLAM by generalizing it to wide-baseline, multi-fisheye camera systems, with optional visual–inertial integration. It fuses dense intra- and inter-view residuals with inertial pre-integration factors within a unified optimization graph. The system further leverages the large field of view for memory-efficient loop closures via semi-pose-graph BA.

\subsection{3D Gaussian Splatting}

3D Gaussian Splatting (3DGS) represents the scene as a set
$\{(\boldsymbol{\mu}_{i},\Sigma_{i},\alpha_{i},\mathbf{c}_{i})\}_{i=1}^{M}$
of $M$ anisotropic Gaussians, where each Gaussian is defined by its mean $\boldsymbol{\mu}_i \in \mathbb{R}^3$, covariance $\Sigma_i \in \mathbb{R}^{3 \times 3}$, opacity $\alpha_i\in\mathbb{R}$, and RGB color $\mathbf{c}_i\in\mathbb{R}^3$. Under a camera pose $\mathbf{T}_i \in \mathrm{SE(3)}$, a 3D Gaussian is projected into an elliptical footprint on the image plane as
\begin{equation}
  \boldsymbol{\mu}'_{i}= \pi(\mathbf{T}_i\boldsymbol{\mu}_{i}),\qquad
  \Sigma'_{i}=J\,R\,\Sigma_{i}R^{\!\top}\!J^{\top},
  \label{eq:gauss_proj}
\end{equation}
where $R$ is the rotational component of $\mathbf{T}$ and $J$ denotes the Jacobian of the perspective projection \cite{kerbl20233d}. The resulting splats are composited in a front-to-back order using $\alpha$-blending to produce color and depth images as
\begin{equation}
  \begin{aligned}
    &\hat{C}(\mathbf{p}) = \sum_{i\in\mathcal{N}_{\mathbf{p}}}
      \mathbf{c}_{i}\alpha_{i}
      \prod_{j<i}\bigl(1-\alpha_{j}\bigr), \\
    &\hat{D}(\mathbf{p}) = \sum_{i\in\mathcal{N}_{\mathbf{p}}}
      d_{i}\alpha_{i}
      \prod_{j<i}\bigl(1-\alpha_{j}\bigr).
  \end{aligned}
  \label{eq:gauss_blend}
\end{equation}
where $\mathcal{N}_{\mathbf{p}}$ denotes the set of Gaussians intersecting the ray. Here, $c_i$ and $d_i$ are the color and depth of the $i$-th Gaussian, respectively, and $\alpha_i$ represents its contribution to pixel translucency, obtained from the Gaussian’s opacity at the ray Gaussian intersection. The projection and blending operations in Equations \eqref{eq:gauss_proj} and \eqref{eq:gauss_blend} are fully differentiable, allowing gradients to be backpropagated with respect to both the camera pose $\mathbf{T}$ and all Gaussian parameters.

\section{Method}
\label{sec:method}

This section presents MCGS-SLAM, a dense multi-camera SLAM pipeline that integrates learning-based tracking with a differentiable 3D Gaussian map representation. An overview of the system architecture is shown in Figure~\ref{fig:method_overview}. The pipeline operates in two stages. In the online tracking stage, the system estimates the camera rig's trajectory in real time, resolves the scale ambiguity associated with monocular priors, and performs MCBA to jointly optimize per-view depths and poses. Refined keyframes are incrementally fused into a global 3DGS map. 
In the optional offline refinement stage, all rig poses and Gaussian parameters are jointly optimized to enforce global consistency and further improve the geometric and photometric fidelity of the reconstruction.

\subsection{Online Multi-Camera Tracking}

\subsubsection{Key-Frame Selection, Depth and Normal Estimation}
For each synchronized RGB frame, we compute the average Recurrent Field Transform (RFT) flow relative to the current reference keyframe. If the flow magnitude exceeds a threshold, the frame is promoted to a multi-camera keyframe, $K_t := \{{I_t, d_t^{\text{+}}, \mathbf{n}_t^{\text{+}}}\}$, where $d_t^{\text{+}}$ and $\mathbf{n}_t^{\text{+}}$ denote the per-pixel depth and surface normal maps. These are obtained from Metric3Dv2~\cite{hu2024metric3d}, which also allows our system to support RGB-D input. Keyframes are stored in a shared buffer accessible to both tracking and mapping threads. Although the depths from Metric3Dv2 are metric, they often suffer from noise, bias, and inconsistent scaling across viewpoints, leading to misaligned poses and depths. Our proposed MCBA module corrects this by jointly refining poses and depths, enforcing geometric consistency and scale alignment across the rig.

\subsubsection{Joint Depth and Scale Alignment (JDSA)}
Depth maps predicted for each RGB camera are only defined up to an unknown, spatially varying scale. To compensate for this ambiguity, \cite{zhang2023hi} introduce a learnable $m \times n$ scale grid $\mathbf{s}_t$ for each key-frame. This grid is bilinearly interpolated to yield a per-pixel scale factor $B_t(\mathbf{p}, \mathbf{s}_t)$, which relates the predicted and optimized depths as $\tilde{d}_{t}(\mathbf{p}) = d^{\text{+}}_{t}(\mathbf{p}) \cdot B_t(\mathbf{p}, \mathbf{s}_t)$, where $d^{\text{+}}_t$ denotes the monocular depth map and $\tilde{d}_t$ the rescaled depth used during optimization. However, directly coupling the scale factors with bundle adjustment, by jointly optimizing camera poses, depths, and scale coefficients, has been shown to cause unstable convergence and scale drift \cite{zhang2024hi}. To mitigate this, we adopt the Joint Depth and Scale Alignment (JDSA) formulation proposed in \cite{zhang2024hi}, which introduces a dedicated loss function as
\begin{align}
\mathop{\arg\min}_{\mathbf{s}, \mathbf{d}} \quad 
& \sum_{(i,j)\in\mathcal{E}} \left\| \tilde{\mathbf{p}}_{ij} - \Pi\left( \mathbf{T}_{ij} \Pi^{-1}(\mathbf{p}_i, \mathbf{d}_i) \right) \right\|^2_{\omega_{ij}} \notag + \\
& \ \: \sum_{i\in\mathcal{V}} \; \; \left\| \tilde{\mathbf{d}}_i \cdot B_i(\mathbf{p}_i, \mathbf{s}_i) - \mathbf{d}_i \right\|^2,
\label{eq:jdsa}
\end{align}
where the node set $\mathcal{V}$ consists of keyframes, each associated with a pose $T \in SE(3)$ and an estimated depth map $d$. The edge set $\mathcal{E}$ connects keyframes that exhibit sufficient overlap, as determined by their optical flow correspondences. The first term enforces multi-view photometric and geometric consistency, and the second term aligns scaled depths to the optimized depths. By interleaving JDSA with local multi-camera bundle adjustment, our system achieves stable scale calibration and improved depth initialization.

\subsubsection{Multi-Camera Bundle Adjustment (MCBA)}

To jointly optimize camera poses and dense depth maps, we minimize a weighted photometric reprojection loss over both temporal and cross-view image pairs. Specifically, for each valid correspondence between a source view \( (i, C_i) \) and a target view \( (j, C_j) \), we define the following objective:
\begin{equation}
\mathop{\arg\min}_{\mathbf{T}, \mathbf{d}} \sum_{(i, j) \in \mathcal{E}} \left\| \tilde{\mathbf{p}}_{ij} - \Pi_{C_j}\left( \hat{\mathbf{T}}_{ij} \cdot \Pi_{C_i}^{-1}(\mathbf{p}_i, d_i) \right) \right\|_{w_{ij}}^2,
\label{eq:reproj}
\end{equation}
where \( \mathbf{T} \in \mathrm{SE}(3) \) denotes the body pose, and \( d_i \) is the estimated inverse depth parametrization in view \( (i, C_i) \). The function \( \Pi_{C_i}^{-1}(\cdot) \) back-projects the pixel using the intrinsics of camera \( C_i \), while \( \Pi_{C_j}(\cdot) \) reprojects it into the target view. The norm $\|\cdot\|_{w_{ij}}$ incorporates a confidence $w_{ij}$ per pixel predicted by the RFT module in multi-camera settings. The transformation \( \hat{\mathbf{T}}_{ij} \) maps 3D points from the source to the target camera frame, and is defined differently based on the type of correspondence as
\begin{itemize}
  \item \textbf{Temporal pairs} (i.e., same camera across time):
  \begin{equation}
  \hat{\mathbf{T}}_{ij} = \mathbf{T}_{C}^{B} \, \mathbf{T}_j^{-1} \, \mathbf{T}_i \, {\mathbf{T}_{C}^{B}}^{-1},
  \end{equation}
  where \( \mathbf{T}_{C}^{B} \) is the known extrinsic between the camera frame $C$ and the body frame $B$.
  
  \item \textbf{Cross-view pairs} (i.e., different cameras at the same timestamp):
  \begin{equation}
  \hat{\mathbf{T}}_{ij} = \mathbf{T}_{C_i C_j},
  \end{equation}
  which is the pre-calibrated extrinsic between camera \( C_i \) and camera \( C_j \).
\end{itemize}
This unified formulation allows for simultaneous optimization over both time-varying motion and multi-camera geometry in a single bundle adjustment framework. The resulting non-linear least-squares problem is solved via a damped Gauss–Newton method, yielding a block-structured linear system of the form as
\begin{equation}
\begin{bmatrix} \mathbf{B} & \mathbf{E}\\
                \mathbf{E}^{\top} & \mathbf{C}
\end{bmatrix}
\begin{bmatrix}\Delta \boldsymbol{\xi}\\ \Delta \mathbf{d}\end{bmatrix} =
\begin{bmatrix}\mathbf{v}\\ \mathbf{w}\end{bmatrix},
\label{eq:gauss_newton}
\end{equation}
where \( \Delta \boldsymbol{\xi} \in \mathbb{R}^6 \) represents the pose update in the Lie algebra of SE(3), applied via \( \Delta \mathbf{T} = \exp(\Delta \boldsymbol{\xi}) \) as in \cite{teed2021droid}. Matrices \( \mathbf{B} \), \( \mathbf{C} \), and \( \mathbf{E} \) correspond to the Hessian blocks with respect to pose, depth, and their coupling terms, while \( \mathbf{v} \) and \( \mathbf{w} \) are the respective residual gradients. Since the pose block \( \mathbf{B} \) is typically much smaller than the depth block \( \mathbf{C} \), we solve the system efficiently using the Schur complement. The pose update is obtained via
\begin{equation}
\begin{aligned}
\Delta \boldsymbol{\xi} &= \left[\mathbf{B} - \mathbf{E}\mathbf{C}^{-1}\mathbf{E}^\top \right]^{-1} \left(\mathbf{v} - \mathbf{E}\mathbf{C}^{-1}\mathbf{w}\right), \\
\Delta \mathbf{d} &= \mathbf{C}^{-1} \left(\mathbf{w} - \mathbf{E}^\top \Delta \boldsymbol{\xi} \right).
\end{aligned}
\end{equation}
In the implementation, the depth Hessian \( \mathbf{C} \) is diagonal and thus admits a cheap closed-form inverse \( \mathbf{C}^{-1} = 1 / \mathbf{C} \).

\subsection{Multi-Camera Gaussian Mapping}

\subsubsection{Gaussian Initialization and Maintenance}
After each MCBA and JDSA update, we back-project the depth map of the latest keyframe \( K_t \) into 3D space to initialize new Gaussian primitives. For each valid pixel \( \mathbf{p} \), a Gaussian \( g_i \) is created with mean \( \boldsymbol{\mu}_i \in \mathbb{R}^3 \) corresponding to the back-projected 3D point, and covariance \( \Sigma_i \in \mathbb{R}^{3 \times 3} \) estimated from the average distance to its three nearest neighbors. To keep the map compact yet expressive, the system alternates every few iterations between two complementary operations: \textbf{(1) densification}, which adds Gaussians at previously unobserved pixels to grow underrepresented regions; and \textbf{(2) pruning}, which removes nearly transparent Gaussians to reduce redundancy and computational overhead.

\subsubsection{Differentiable Rasterization and Losses}
We follow \cite{zhang2024hi} and avoid depth bias by analytically intersecting each viewing ray with the ellipsoidal surface defined by the anisotropic Gaussian, yielding a more accurate intersection depth. Each Gaussian is jointly optimized through a multi-term loss function per keyframe \( K_t \) as 
\begin{align}
\mathcal{L} =&\; \lambda_c \left\| \hat{C}_t - I_{t} \right\|_2
+ \lambda_d \left\| \hat{D}_t - d_{t} \right\|_2 \notag \\
&+ \lambda_n \left\| 1 - \left\langle \hat{\mathbf{n}}_t, \mathbf{n}^{\text{pri}}_t \right\rangle \right\|_2
+ \lambda_s \left\| \mathbf{s}_t - \bar{\mathbf{s}} \right\|_2,
\label{eq:multi_loss}
\end{align}
where \( \hat{C}_t \) and \( \hat{D}_t \) denote the rendered color and depth from the viewpoint of the MCBA-refined camera pose, \( d_t \) is the depth refined via MCBA and JDSA, \( \hat{\mathbf{n}}_t \) and \( \mathbf{n}_t^{\text{+}} \) represent the rendered and estimated surface normals by Metric3Dv2, and \( \mathbf{s}_t \) and \( \bar{\mathbf{s}} \) denote the current and average scale of the corresponding Gaussian ellipsoids, respectively. Optimization is performed using the optimizer for a fixed number of iterations per keyframe.

\subsubsection{Pose-Consistent Gaussian Updates}
When the pose of a keyframe \( K_t \) is updated via MCBA or loop closure by a relative transform \( \Delta \mathbf{T}_t \in \text{SE}(3) \), we propagate the update to all Gaussians \( g_i \) anchored in that frame as
\begin{equation}
\boldsymbol{\mu}_i \leftarrow \Delta \mathbf{T}_t \cdot \boldsymbol{\mu}_i, \quad
\Sigma_i \leftarrow \mathbf{R}(\Delta \mathbf{T}_t) \cdot \Sigma_i \cdot \mathbf{R}^{\top}(\Delta \mathbf{T}_t),
\label{eq:gaussian_transform}
\end{equation}
where \( \mathbf{R}(\cdot) \) extracts the rotational component of \( \Delta \mathbf{T}_t \). If scale changes are introduced via scale updates, we additionally rescale the ellipsoids as
\begin{equation}
\mathbf{s}_i \leftarrow s_t \cdot \mathbf{s}_i.
\end{equation}
This deformation ensures consistency of the 3D map without requiring re-initialization or re-rendering, enabling efficient and flexible map maintenance.

\subsection{Offline Global Refinement}

after the real-time pipeline finishes, we apply two global refinement stages to enhance the consistency and overall quality of the reconstruction.

\subsubsection{Global Bundle Adjustment}

All keyframes that includes synthetically inserted views, are jointly optimized via global bundle adjustment. The optimization minimizes both photometric and geometric residuals across all overlapping image pairs, refining the camera poses and improving the consistency of the reconstructed scene geometry.

\subsubsection{Joint Pose and 3DGS Map Refinement}

In the final stage, we jointly optimize all 3D Gaussian parameters  
\(\boldsymbol{\Theta} := \left\{ \boldsymbol{\mu}, \Sigma, \alpha, \mathbf{c} \right\}\),  
along with per-frame exposure matrices \( \mathbf{A}_t \) and camera poses \( \mathbf{T}_t \). Gradients are backpropagated through the differentiable rasterization pipeline to minimize a weighted combination of photometric, depth, normal, and scale regularization losses. This optimization stage effectively reduces global drift and improves both the geometric accuracy and photometric consistency of the final reconstruction.

\section{Results}
\label{sec:results}

\subsection{Datasets, Metrics, and Protocol}
We evaluated MCGS-SLAM on both real-world and synthetic datasets. For real-world experiments, we employ the \textbf{Waymo Open Dataset}~\cite{sun2020scalability}, which provides urban driving sequences with five synchronized wide-angle roof cameras. We select three of them, as this already ensures a sufficiently wide front-facing field of view while keeping GPU memory usage manageable. We further use the \textbf{Oxford Spires Dataset}\cite{tao2025spires}, which contains large-scale Oxford landmarks recorded by three fisheye cameras with LiDAR/IMU ground truth. For synthetic evaluation, we adopt the \textbf{AirSim}~\cite{airsim2017fsr} simulator with three photorealistic UE5 environments, captured using a four-camera aircraft rig in the simulation setting. Reconstruction quality is quantified using standard image-based metrics: PSNR ($\uparrow$), SSIM ($\uparrow$) and LPIPS ($\downarrow$) - computed over all keyframes after mapping. The trajectory accuracy is measured by the absolute trajectory error (ATE, meters; $\downarrow$) after Sim(3)-alignment with the ground truth. For better readability, the result tables highlight the top three results with \colorbox{colorFst}{\textbf{first}}, \colorbox{colorSnd}{second}, and \colorbox{colorTrd}{third}.

\subsection{Rendering Results Study}

\begin{figure*}[t]
  \centering
  \includegraphics[scale=0.365]{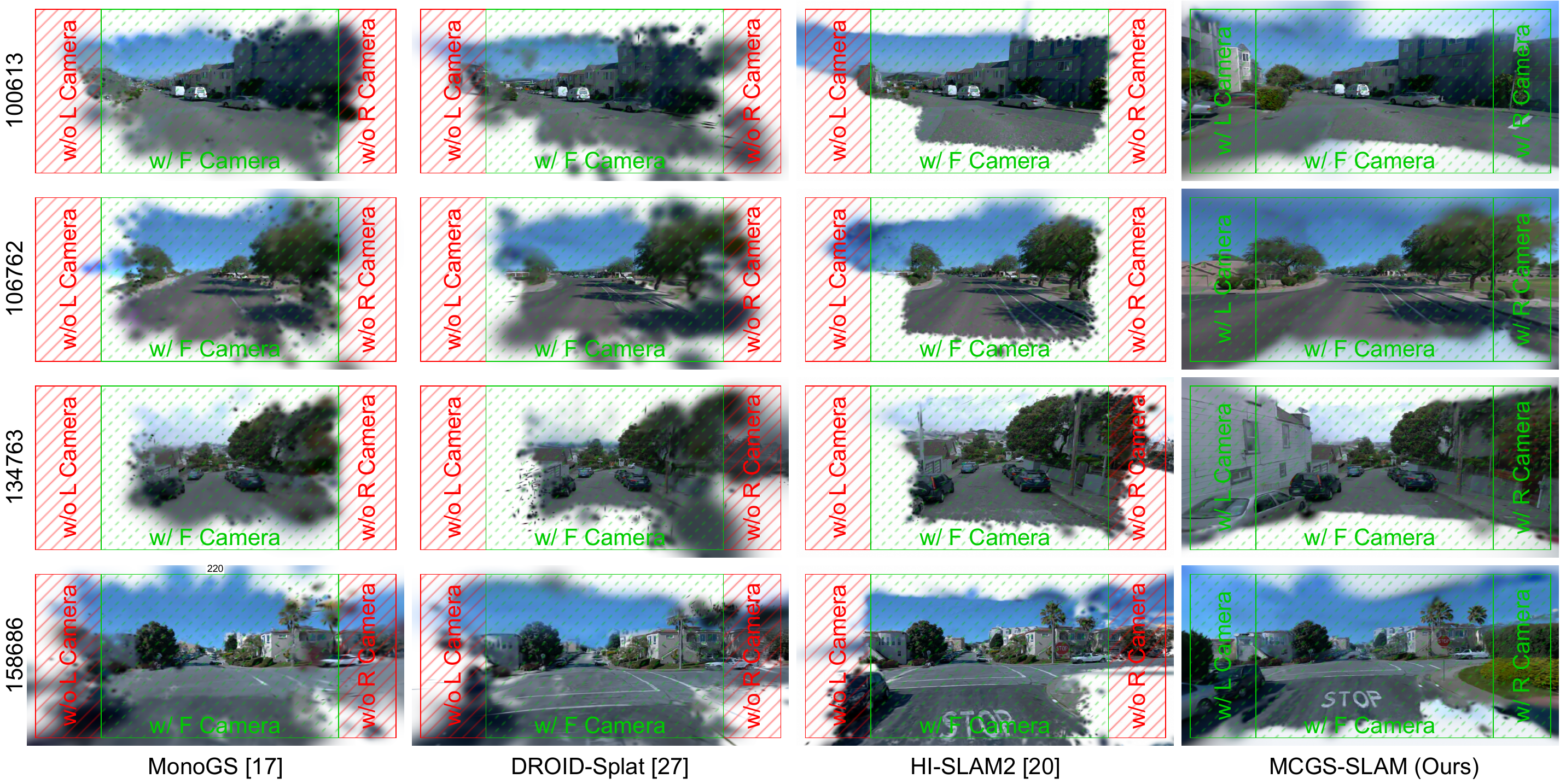}
  \vspace{-0.1cm} 
  \caption{Qualitative results on the Waymo dataset~\cite{sun2020scalability} (\textbf{Real-World Dataset}). MCGS-SLAM reconstructs urban scenes with higher fidelity and completeness, preserving structural details and textures that are often missed by monocular methods.}
  \label{fig:compared_waymo}
  \vspace{-0.2cm}  
\end{figure*}

\begin{figure*}[tb]
  \vspace{0.1cm}
  \centering
  \includegraphics[scale=0.365]{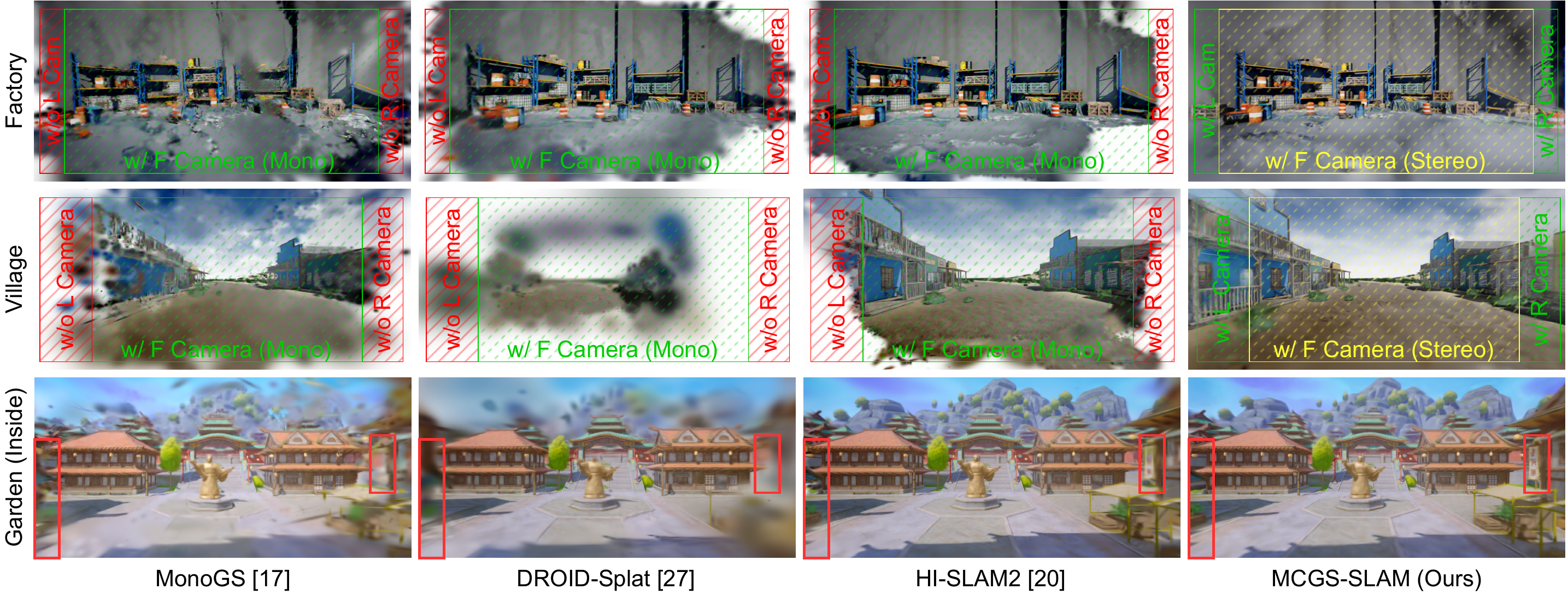}
  \caption{MCGS-SLAM produces faithful and complete reconstructions on AirSim~\cite{airsim2017fsr} (\textbf{Synthetic Dataset}).}
  \label{fig:compared_airsim}
  \vspace{-0.2cm}  
\end{figure*}

Tables~\ref{tab:waymo_recon_results} and \ref{tab:airsim_recon_results} present quantitative appearance metrics, while Figures~\ref{fig:compared_waymo} and \ref{fig:compared_airsim} show qualitative reconstruction results in different Waymo and AirSim environments. On the four held-out urban sequences from the Waymo dataset, MCGS-SLAM consistently ranks among the top two performers, demonstrating strong photometric fidelity and perceptual quality. In contrast, competing methods report inferior LPIPS values and fail to reconstruct critical side-view structures, such as alley facades, that are clearly recovered by MCGS-SLAM (see Fig.\ref{fig:compared_waymo}). This advantage stems from the wide field of view (FoV) provided by the multi-camera rig (Fig.~\ref{fig:waymo_dataset}), which enables MCGS-SLAM to resolve occluded elements such as building corners and overhead traffic lights. Furthermore, the resulting 3D maps exhibit substantially fewer floating artifacts, highlighting the effectiveness of cross-view depth consistency enforced by our MCBA and JDSA modules. Although GLORIE-SLAM and DROID-Splat occasionally reconstruct sharper specular surfaces, their limited spatial coverage leads to incomplete scene geometry. Overall, MCGS-SLAM achieves a better balance between reconstruction fidelity and spatial completeness, making it particularly well suited for complex urban environments.

\begin{table*}[tb]
\vspace{0.1cm} 
\centering
\begin{adjustbox}{max width=\textwidth}
\begin{tabular}{lcccccccccc}
\toprule
\textbf{Method} & \,\,\textbf{Metric}\,\, 
& \,\,100613\,\, & \,\,132384\,\, & \,\,134763\,\, & \,\,152706\,\, & \,\,158686\,\, & \,\,153495\,\, & \,\,106762\,\, & \,\,163453\,\, & \,\,Avg.\,\, \\
\midrule
\multirow{3}{*}{NICER-SLAM~\cite{zhu2024nicer}} 
  & PSNR\;$\uparrow$ & 12.91 & 15.48 & 8.79  & 11.32 & 11.68 & 13.25 & 13.09 & 16.41 & 12.87 \\
  & SSIM\;\;$\uparrow$ & 0.498 & 0.775 & 0.330 & 0.611 & 0.438 & 0.541 & 0.587 & 0.712 & 0.562 \\
  & LPIPS\;$\downarrow$ & 0.695 & 0.518 & 0.791 & 0.754 & 0.686 & 0.691 & 0.626 & 0.657 & 0.677 \\
\midrule
\multirow{3}{*}{GLORIE-SLAM~\cite{zhang2024glorie}} 
  & PSNR\;$\uparrow$ & \cellcolor{colorTrd}25.78 & \cellcolor{colorSnd}25.52 & \cellcolor{colorTrd}25.71 & \cellcolor{colorTrd}24.90 & \cellcolor{colorSnd}25.09 & \cellcolor{colorSnd}23.79 & \cellcolor{colorSnd}27.35 & \cellcolor{colorSnd}23.72 & \cellcolor{colorSnd}25.23 \\
  & SSIM\;\;$\uparrow$ & \cellcolor{colorFst}\textbf{0.916} & \cellcolor{colorFst}\textbf{0.902} & \cellcolor{colorFst}\textbf{0.883} & \cellcolor{colorFst}\textbf{0.878} & \cellcolor{colorFst}\textbf{0.908} & \cellcolor{colorFst}\textbf{0.891} & \cellcolor{colorFst}\textbf{0.918} & \cellcolor{colorFst}\textbf{0.903} & \cellcolor{colorFst}\textbf{0.900} \\
  & LPIPS\;$\downarrow$ & \cellcolor{colorTrd}0.282 & \cellcolor{colorSnd}0.287 & \cellcolor{colorSnd}0.365 & \cellcolor{colorSnd}0.338 & \cellcolor{colorFst}\textbf{0.291} & \cellcolor{colorSnd}0.309 & \cellcolor{colorSnd}0.272 & \cellcolor{colorSnd}0.279 & \cellcolor{colorSnd}0.303 \\
\midrule
\multirow{3}{*}{MonoGS~\cite{matsuki2024gaussian}} 
  & PSNR\;$\uparrow$ & 20.58 & 23.53 & 21.41 & 22.34 & 21.87 & 21.08 & 22.31 & 19.41 & \cellcolor{colorTrd}21.57 \\
  & SSIM\;\;$\uparrow$ & 0.674 & \cellcolor{colorSnd}0.862 & 0.620 & \cellcolor{colorTrd}0.784 & \cellcolor{colorTrd}0.684 & \cellcolor{colorTrd}0.772 & 0.741 & 0.753 & \cellcolor{colorTrd}0.737 \\
  & LPIPS\;$\downarrow$ & 0.607 & 0.421 & 0.625 & 0.641 & 0.514 & 0.646 & 0.503 & 0.657 & 0.577 \\
\midrule
\multirow{3}{*}{DROID-Splat~\cite{homeyer2024droid}} 
  & PSNR\;$\uparrow$ & \cellcolor{colorSnd}26.77 & \cellcolor{colorTrd}25.02 & \cellcolor{colorSnd}26.20 & \cellcolor{colorSnd}25.92 & \cellcolor{colorFst}\textbf{26.81} & \cellcolor{colorTrd}24.01 & \cellcolor{colorTrd}27.21 & \cellcolor{colorTrd}23.02 & 25.62 \\
  & SSIM\;\;$\uparrow$ & \cellcolor{colorTrd}0.829 & 0.823 & \cellcolor{colorTrd}0.792 & 0.782 & \cellcolor{colorSnd}0.850 & 0.748 & \cellcolor{colorSnd}0.864 & 0.720 & \cellcolor{colorSnd}0.801 \\
  & LPIPS\;$\downarrow$ & \cellcolor{colorSnd}0.273 & 0.384 & \cellcolor{colorTrd}0.376 & 0.512 & \cellcolor{colorSnd}0.297 & 0.482 & \cellcolor{colorTrd}0.281 & 0.451 & \cellcolor{colorTrd}0.382 \\
\midrule
\multirow{3}{*}{Photo-SLAM~\cite{huang2024photo}} 
  & PSNR\;$\uparrow$ & 19.03 & 20.49 & 21.28 & 20.84 & 21.44 & 18.14 & 20.26 & 19.12 & 20.08 \\
  & SSIM\;\;$\uparrow$ & 0.640 & 0.824 & 0.624 & 0.759 & 0.674 & 0.726 & 0.712 & \cellcolor{colorTrd}0.758 & 0.715 \\
  & LPIPS\;$\downarrow$ & 0.527 & \cellcolor{colorTrd}0.307 & 0.471 & \cellcolor{colorTrd}0.466 & \cellcolor{colorTrd}0.367 & \cellcolor{colorTrd}0.453 & 0.440 & 0.444 & 0.434 \\
\midrule
\multirow{3}{*}{MCGS-SLAM (Ours)} 
  & PSNR\;$\uparrow$ & \cellcolor{colorFst}\textbf{27.09} & \cellcolor{colorFst}\textbf{26.26} & \cellcolor{colorFst}\textbf{27.20} & \cellcolor{colorFst}\textbf{28.45} & \cellcolor{colorTrd}21.91 & \cellcolor{colorFst}\textbf{26.48} & \cellcolor{colorFst}\textbf{27.70} & \cellcolor{colorFst}\textbf{26.92} & \cellcolor{colorFst}\textbf{26.50} \\
  & SSIM\;\;$\uparrow$ & \cellcolor{colorSnd}0.830 & \cellcolor{colorTrd}0.826 & \cellcolor{colorSnd}0.813 & \cellcolor{colorSnd}0.797 & 0.682 & \cellcolor{colorSnd}0.813 & \cellcolor{colorTrd}0.819 & \cellcolor{colorSnd}0.829 & \cellcolor{colorSnd}0.801 \\
  & LPIPS\;$\downarrow$ & \cellcolor{colorFst}\textbf{0.223} & \cellcolor{colorFst}\textbf{0.284} & \cellcolor{colorFst}\textbf{0.233} & \cellcolor{colorFst}\textbf{0.330} & 0.547 & \cellcolor{colorFst}\textbf{0.231} & \cellcolor{colorFst}\textbf{0.262} & \cellcolor{colorFst}\textbf{0.234} & \cellcolor{colorFst}\textbf{0.293} \\
\bottomrule
\end{tabular}
\end{adjustbox}
\caption{Appearance reconstruction comparison of different methods on 8 scenes of the Waymo dataset~\cite{sun2020scalability} (\textbf{Real-World Dataset}). MCGS-SLAM achieves the best PSNR and LPIPS results, highlighted as \colorbox{colorFst}{\textbf{first}}, \colorbox{colorSnd}{second} and \colorbox{colorTrd}{third}.}
\label{tab:waymo_recon_results}
\vspace{-0.6cm} 
\end{table*}

Similar trends are observed in the AirSim benchmark (Table~\ref{tab:airsim_recon_results} and Fig.~\ref{fig:compared_airsim}), where MCGS-SLAM consistently ranks among the top two methods across all environments. In the low-parallax \textit{Garden} scene, it surpasses all single-camera baselines by approximately 4\,dB PSNR, highlighting the benefit of leveraging complementary viewpoints from a wide-baseline rig. In the \textit{Factory} scene, although Photo-SLAM achieves the highest PSNR and SSIM, MCGS-SLAM yields cleaner and more geometrically consistent reconstructions thanks to dense cross-view constraints and broader visual coverage. The \textit{Village} scene, characterized by abrupt turns and large FoV discontinuities, remains challenging for single-camera baselines (e.g., MonoGS, DROID-Splat), which exhibit holes and blending artifacts. By exploiting multi-view priors and robust depth-scale alignment, MCGS-SLAM reconstructs sharper structures and more complete geometry even under wide-baseline motion.

\begin{table*}[htbp]
\centering
\begin{adjustbox}{max width=\textwidth}
\begin{tabular}{lcccccccccc}
\toprule
\textbf{Method} & \textbf{Metric} & \,\,100613\,\, & \,\,158686\,\, & \,\,132384\,\, & \,\,134763\,\, & \,\,152706\,\, & \,\,153495\,\, & \,\,106762\,\, & \,\,163453\,\, & \,\,\,\,Avg.\,\,\,\, \\
\midrule
\multirow{1}{*}{\,NICER-SLAM~\cite{zhu2024nicer}\,} 
  & \,ATE [m]\;\;$\downarrow$\, & 2.351 & \cellcolor{colorTrd}2.362 & 56.363 & 2.642 & 19.409 & 19.782 & 1.634 & 14.708 & 14.906 \\
\multirow{1}{*}{\,MonoGS~\cite{matsuki2024gaussian}\,} 
  & \,ATE [m]\;\;$\downarrow$\, & 10.727 & 10.101 & 12.033 & 3.394 & 9.073 & \cellcolor{colorTrd}1.628 & 19.532 & 9.189 & 9.459 \\
\multirow{1}{*}{\,Splat-SLAM~\cite{sandstrom2025splat}\,} 
  & \,ATE [m]\;\;$\downarrow$\, & \cellcolor{colorTrd}0.802 & 2.575 & \cellcolor{colorSnd}1.133 & \cellcolor{colorTrd}1.625 & \cellcolor{colorSnd}1.092 & 2.572 & \cellcolor{colorFst}\textbf{1.973} & \cellcolor{colorTrd}3.115 & \cellcolor{colorTrd}1.861 \\
\multirow{1}{*}{\,HI-SLAM2~\cite{zhang2024hi}\,} 
  & \,ATE [m]\;\;$\downarrow$\, & \cellcolor{colorSnd}0.790 & \cellcolor{colorSnd}1.782 & \cellcolor{colorFst}\textbf{0.888} & \cellcolor{colorSnd}1.281 & \cellcolor{colorFst}\textbf{0.964} & \cellcolor{colorSnd}1.389 & \cellcolor{colorSnd}2.132 & \cellcolor{colorSnd}2.558 & \cellcolor{colorSnd}1.473 \\
\multirow{1}{*}{\,MCGS-SLAM (Ours)\,} 
  & \,ATE [m]\;\;$\downarrow$\, & \cellcolor{colorFst}\textbf{0.398} & \cellcolor{colorFst}\textbf{0.612} & \cellcolor{colorTrd}1.242 & \cellcolor{colorFst}\textbf{1.107} & \cellcolor{colorTrd}2.554 & \cellcolor{colorFst}\textbf{1.180} & \cellcolor{colorTrd}2.366 & \cellcolor{colorFst}\textbf{0.927} & \cellcolor{colorFst}\textbf{1.298} \\
\bottomrule
\end{tabular}
\end{adjustbox}
\caption{Quantitative comparison of tracking accuracy (ATE RMSE) across different methods and scenes on the Waymo dataset~\cite{sun2020scalability}. MCGS-SLAM yields the best average results. Best results are highlighted as \colorbox{colorFst}{\textbf{first}}, \colorbox{colorSnd}{second} and \colorbox{colorTrd}{third}.}
\label{tab:waymo_track_results}
\vspace{-0.4cm}
\end{table*}

\begin{table}[tb]
\centering
\begin{adjustbox}{max width=\textwidth}
\begin{tabular}{l|c|cccc}
\toprule
\textbf{Method} & \textbf{Metric} 
& Garden & Factory & Village & \,\,Avg.\,\, \\
\midrule
\multirow{3}{*}{\makecell[l]{NICER- \\ SLAM~\cite{zhu2024nicer}}} 
  & PSNR\;$\uparrow$ & 12.30 & 9.84 & 11.18 & 11.11\\
  & SSIM\;\;$\uparrow$ & 0.450 & 0.332 & 0.504 & 0.429 \\
  & LPIPS\;$\downarrow$ & 0.801 & 0.690 & 0.653 & 0.715 \\
\midrule
\multirow{3}{*}{\makecell[l]{GLORIE- \\ SLAM~\cite{zhang2024glorie}}} 
  & PSNR\;$\uparrow$ & 24.50 & 23.39 & 17.56 & 21.82 \\
  & SSIM\;\;$\uparrow$ & \cellcolor{colorSnd}0.849 & 0.888 & 0.494 & 0.744 \\
  & LPIPS\;$\downarrow$ & 0.351 & 0.346 & 0.712 & 0.470 \\
\midrule
\multirow{3}{*}{MonoGS~\cite{matsuki2024gaussian}} 
  & PSNR\;$\uparrow$ & \cellcolor{colorSnd}25.59 & 21.45 & \cellcolor{colorTrd}21.39 & \cellcolor{colorTrd}22.81 \\
  & SSIM\;\;$\uparrow$ & 0.766 & 0.760 & \cellcolor{colorTrd}0.689 & 0.738 \\
  & LPIPS\;$\downarrow$ & 0.258 & 0.175 & \cellcolor{colorTrd}0.444 & \cellcolor{colorTrd}0.292 \\
\midrule
\multirow{3}{*}{\makecell[l]{DROID- \\ Splat~\cite{homeyer2024droid}}} 
  & PSNR\;$\uparrow$ & 24.12 & \cellcolor{colorTrd}26.50 & 17.25 & 22.62 \\
  & SSIM\;\;$\uparrow$ & \cellcolor{colorTrd}0.822 & \cellcolor{colorTrd}0.898 & 0.669 & \cellcolor{colorTrd}0.796 \\
  & LPIPS\;$\downarrow$ & \cellcolor{colorTrd}0.242 & \cellcolor{colorTrd}0.107 & 0.652 & 0.334 \\
\midrule
\multirow{3}{*}{\makecell[l]{Photo- \\ SLAM~\cite{huang2024photo}}} 
  & PSNR\;$\uparrow$ & \cellcolor{colorTrd}25.47 & \cellcolor{colorFst}\textbf{28.38} & \cellcolor{colorSnd}26.77 & \cellcolor{colorSnd}26.87 \\
  & SSIM\;\;$\uparrow$ & 0.775 & \cellcolor{colorSnd}0.923 & \cellcolor{colorSnd}0.805 & \cellcolor{colorSnd}0.834 \\
  & LPIPS\;$\downarrow$ & \cellcolor{colorSnd}0.156 & \cellcolor{colorFst}\textbf{0.041} & \cellcolor{colorFst}\textbf{0.205} & \cellcolor{colorFst}\textbf{0.134} \\
\midrule
\multirow{3}{*}{\makecell[l]{MCGS- \\ SLAM (Ours)}} 
  & PSNR\;$\uparrow$ & \cellcolor{colorFst}\textbf{29.36} & \cellcolor{colorSnd}28.37 & \cellcolor{colorFst}\textbf{28.10} & \cellcolor{colorFst}\textbf{28.64} \\
  & SSIM\;\;$\uparrow$ & \cellcolor{colorFst}\textbf{0.879} & \cellcolor{colorFst}\textbf{0.924} & \cellcolor{colorFst}\textbf{0.853} & \cellcolor{colorFst}\textbf{0.885} \\
  & LPIPS\;$\downarrow$ & \cellcolor{colorFst}\textbf{0.126} & \cellcolor{colorSnd}0.083 & \cellcolor{colorSnd}0.219 & \cellcolor{colorSnd}0.143 \\
\bottomrule
\end{tabular}
\end{adjustbox}
\caption{Quantitative comparison of appearance reconstructions of different methods on 3 scenes of the AirSim dataset~\cite{airsim2017fsr} (\textbf{Synthetic Dataset}). 
Best results are highlighted as \colorbox{colorFst}{\textbf{first}}, \colorbox{colorSnd}{second} and \colorbox{colorTrd}{third}.}
\label{tab:airsim_recon_results}
\vspace{-0.8cm}
\end{table}

\subsection{Tracking Results Study}
Tables~\ref{tab:waymo_track_results} and \ref{tab:oxford_spire_track_results} summarize the quantitative tracking accuracy on diverse real-world datasets. These Waymo sequences use original images without distortion correction, providing a more challenging and realistic setting to evaluate tracking robustness in autonomous driving conditions. For the Oxford Spires dataset~\cite{tao2025spires}, the original fisheye images were undistorted to fit the pinhole camera model, and sequences with severe distortion were excluded. MCGS-SLAM achieves the lowest average ATE and ranks first in five of eight Waymo sequences, demonstrating strong robustness to wide baselines and complex environments. It maintains low drift even in difficult cases such as 100613 and 106762, where methods like MonoGS show large trajectory errors. To account for the scene-dependent behavior of the JDSA module, which improves metric-scale consistency, but can occasionally increase drift, we evaluated both configurations and reported the better result. The performance gains mainly stem from the joint optimization of inter-camera depth and pose in the MCBA module, supported by effective scale alignment via JDSA. Although HI-SLAM2 and Splat-SLAM perform competitively in terms of ATE, their monocular design leads to greater drift in long or wide-baseline sequences.

Similar trends appear in the Oxford Spires dataset, which features complex large-scale outdoor scenes with frequent occlusions and strong parallax. MCGS-SLAM again delivers superior performance, achieving the lowest average ATE and outperforming all baselines by a significant margin. In contrast, MonoGS fails on several sequences, leading to heavily degraded ATE values, while HI-SLAM2 and Splat-SLAM suffer from scale ambiguity and tracking discontinuities. The ability of MCGS-SLAM to leverage multi-view redundancy and consistently recover occluded structures from multiple viewpoints proves essential in these challenging, large-scale environments. Overall, the results underscore the robustness and accuracy of our multi-camera framework, which achieves drift-resilient tracking and significantly outperforms prior monocular and single-camera systems.

\begin{table}[tb]
\centering
\begin{adjustbox}{max width=\textwidth}
\begin{tabular}{l|cccc}
\toprule
\textbf{Method} & Library & Palace & College & Observatory \\
\midrule
\multirow{1}{*}{NICER-SLAM~\cite{zhu2024nicer}}
   & 77.593 & 41.593 & 24.580 & 23.621 \\
\multirow{1}{*}{MonoGS~\cite{matsuki2024gaussian}} 
  & FAILED & \cellcolor{colorSnd}29.451 & 30.794 & \cellcolor{colorTrd}11.814 \\
\multirow{1}{*}{Splat-SLAM~\cite{sandstrom2025splat}}
  & \cellcolor{colorTrd}11.890 & 37.853 & \cellcolor{colorTrd}5.756 & 19.727 \\
\multirow{1}{*}{HI-SLAM2~\cite{zhang2024hi}} 
  & \cellcolor{colorSnd}9.001 & \cellcolor{colorTrd}31.601 & \cellcolor{colorSnd}1.694 & \cellcolor{colorFst}\textbf{0.262} \\
\multirow{1}{*}{MCGS-SLAM (Ours)} 
  & \cellcolor{colorFst}\textbf{7.665} & \cellcolor{colorFst}\textbf{3.391} & \cellcolor{colorFst}\textbf{1.551} & \cellcolor{colorSnd}0.924 \\
\bottomrule
\end{tabular}
\end{adjustbox}
\caption{Quantitative comparison of tracking accuracy (ATE RMSE) across different methods and scenes on the Oxford Spires Dataset (Bodleian Library, Blenheim Palace, Christ Church College, and Observatory Quarter). Best results are highlighted as \colorbox{colorFst}{\textbf{first}}, \colorbox{colorSnd}{second} and \colorbox{colorTrd}{third}.}
\label{tab:oxford_spire_track_results}
\vspace{-0.5cm}
\end{table}

\begin{figure*}[htb]
  \centering
  \includegraphics[scale=0.33]{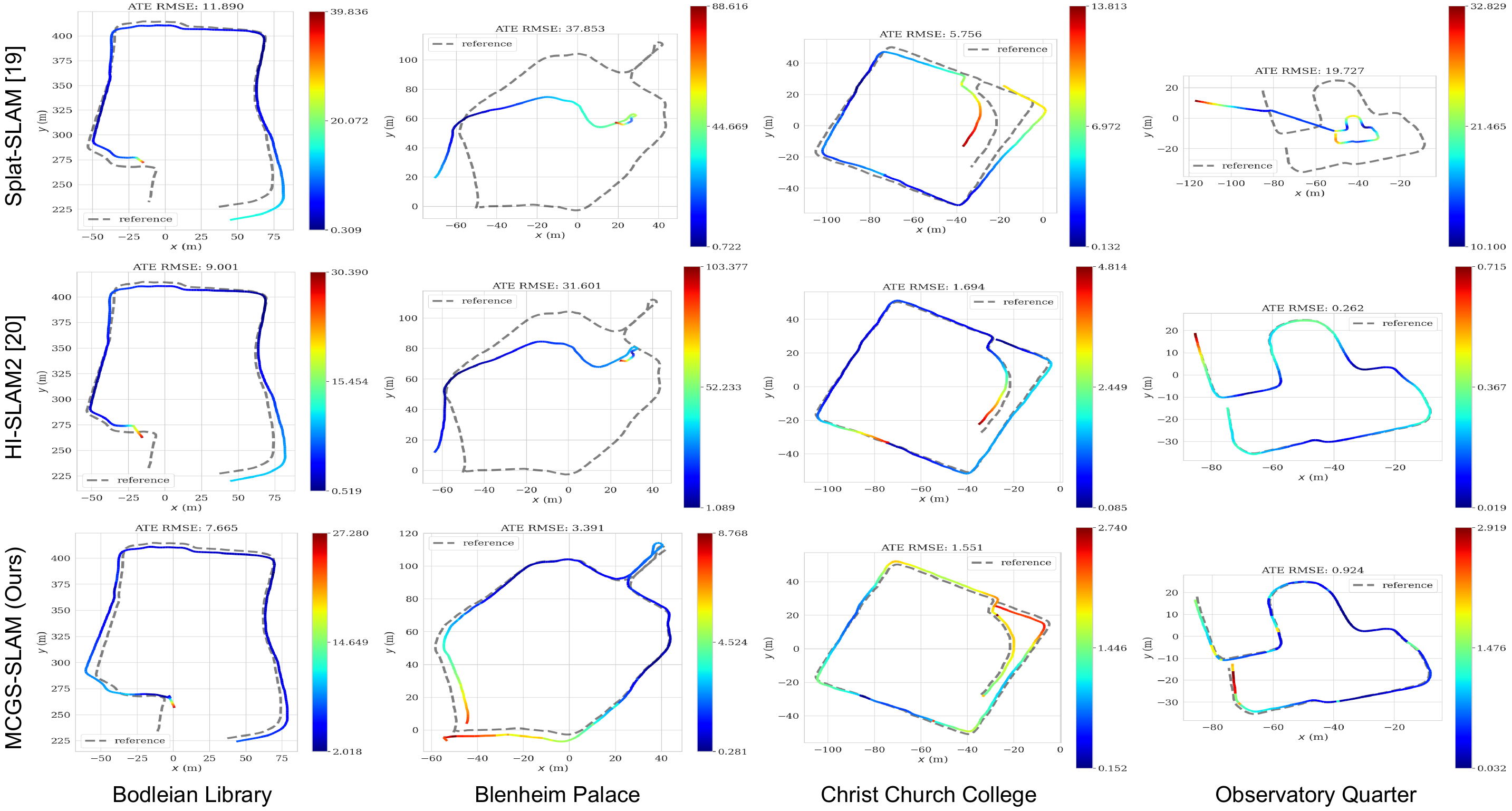}
  \vspace{0.1cm}
  \caption{Tracking performance on the Oxford Spires Dataset \cite{tao2025spires}, evaluated across 4 representative sequences. Ground truth trajectories are compared against Splat-SLAM~\cite{sandstrom2025splat}, HI-SLAM2~\cite{zhang2024hi}, and our MCGS-SLAM. MCGS-SLAM remains closely aligned with ground truth across all sequences, usually achieving the lowest ATE RMSE values and demonstrating the robustness and accuracy of our multi-camera framework in large-scale outdoor environments.}
  \vspace{-0.4cm}
  \label{fig:compared_tracking}
\end{figure*}
%

\subsection{Ablation Study}

Table~\ref{tab:ablation} analyzes the contributions of the Joint Depth–Scale Alignment (JDSA) module and the monocular depth maps predicted by Metric3Dv2~\cite{hu2024metric3d}. Removing both components leads to a notable degradation in performance, with PSNR dropping, SSIM decreasing, and LPIPS increasing substantially. Introducing only the previous depth improves PSNR, but results in subtle double-edge artifacts due to inconsistent estimates of the per-camera scale. In contrast, the full configuration of MCGS-SLAM, with both JDSA and estimated monocular depth, achieves the best scores across all three metrics, justifying its superior photometric accuracy. Although this combination improves reconstruction quality, in a few challenging cases the ATE slightly worsens, likely due to errors in the depth maps, suggesting that stronger predictors could further enhance consistency. Qualitatively, the depth estimates enhance depth initialization, providing a better starting point for optimization in multi-camera configurations. However, without JDSA, inter-camera scale inconsistencies persist, leading to visible artifacts. The JDSA module corrects these inconsistencies by performing per-camera scale alignment and compensates for missing or noisy depth estimates from MCBA. The combined effect of depth initialization and scale-consistent optimization improves depth reliability. Their synergy is key to exploiting the wide field-of-view, enabling cross-view alignment that densifies Gaussians in regions unseen by a single lens.

\begin{table}[tb]
\vspace{0.2cm} 
  \centering
  \setlength{\tabcolsep}{11.5pt}  
  \begin{tabular}{lccc}
    \toprule
    Method & PSNR\,{\small↑} & SSIM\,{\small↑} & LPIPS\,{\small↓} \\
    \midrule
     w/o Depth$^*$ + w/o JDSA & \rd 25.01 & \rd 0.751 & \rd 0.404 \\
     w/ Depth$^*$ + w/o JDSA  & \nd 27.02 & \nd 0.809 & \nd 0.271 \\
     MCGS-SLAM (full)     & \fs 27.17 & \fs 0.816 & \fs 0.262 \\
    \bottomrule
  \end{tabular}
  \vspace{-0.2cm}
    \begin{flushleft}
    \footnotesize
    $^*$From Metric3Dv2~\cite{hu2024metric3d}
    \end{flushleft}
    \vspace{-0.1cm} 
  \caption{Ablation of Joint Depth–Scale Alignment (JDSA) on the Waymo dataset~\cite{sun2020scalability}, averaged over 6 sequences (134763, 106762, 132384, 152706, 153495, and 163453). Best results are highlighted as \colorbox{colorFst}{\textbf{first}}, \colorbox{colorSnd}{second} and \colorbox{colorTrd}{third}.}
  \label{tab:ablation}
  \vspace{-0.5cm}
\end{table}

\section{Conclusion and Future Work}

In this work, we introduced MCGS-SLAM, a fully vision-based SLAM framework that constructs unified 3D Gaussian maps from synchronized multi-camera RGB inputs. By jointly optimizing camera poses and dense depths through Multi-Camera Bundle Adjustment (MCBA) and enforcing inter-camera scale consistency via our proposed Joint Depth–Scale Alignment (JDSA) module, the system achieves real-time, photorealistic, and geometrically consistent reconstructions. MCGS-SLAM performs well in both synthetic and real-world scenarios. Our analysis highlights the critical role of wide-baseline, overlapping views in enhancing scene completeness and robustness, particularly under occlusion and viewpoint discontinuities where monocular systems often fail. Looking ahead, promising directions include integrating inertial or event-based sensing for improved performance in dynamic or low-texture environments, extending the system to support uncalibrated or asynchronous rigs, and further incorporating semantic or instance-level understanding for object-aware mapping.


\bibliography{references}
\bibliographystyle{unsrt}

\end{document}